# ATCSpeechNet: A multilingual end-to-end speech recognition framework for air traffic control systems

Yi Lin, Bo Yang, Linchao Li, Dongyue Guo, Jianwei Zhang, Hu Chen, Yi Zhang

***Abstract*— In this paper, a multilingual end-to-end framework, called as ATCSpeechNet, is proposed to tackle the issue of translating communication speech into human-readable text in air traffic control (ATC) systems. In the proposed framework, we focus on integrating the multilingual automatic speech recognition (ASR) into one model, in which an end-to-end paradigm is developed to convert speech waveform into text directly, without any feature engineering or lexicon. In order to make up for the deficiency of the handcrafted feature engineering caused by ATC challenges, including multilingual, multi-speaker dialog and unstable speech rate, a speech representation learning (SRL) network is proposed to capture robust and discriminative speech representations from the raw wave. The self-supervised training strategy is adopted to optimize the SRL network from unlabeled data, and further to predict the speech features, i.e., wave-to-feature. An end-to-end architecture is improved to complete the ASR task, in which a grapheme-based modeling unit is applied to address the multilingual ASR issue. Facing the problem of small transcribed samples in the ATC domain, an unsupervised approach with mask prediction is applied to pre-train the backbone network of the ASR model on unlabeled data by a feature-to-feature process. Finally, by integrating the SRL with ASR, an end-to-end multilingual ASR framework is formulated in a supervised manner, which is able to translate the raw wave into text in one model, i.e., wave-to-text. Experimental results on the ATCSpeech corpus demonstrate that the proposed approach achieves a high performance with a very small labeled corpus and less resource consumption, only 4.20% label error rate on the 58-hour transcribed corpus. Compared to the baseline model, the proposed approach obtains over 100% relative performance improvement which can be further enhanced with the increasing of the size of the transcribed samples. It is also confirmed that the proposed SRL and training strategies make significant contributions to improve the final performance. In addition, the effectiveness of the proposed framework is also validated on common corpora (AISHELL and Librispeech). More importantly, the proposed multilingual framework not only reduces the system complexity, but also obtains higher accuracy compared to that of the independent monolingual ASR models. The proposed approach can also greatly save the cost of annotating samples, which benefits to advance the ASR technique to industrial applications. Based on the proposed framework, a real-time sensing approach is expected to be implemented to further support ATC-related applications.**

## I. INTRODUCTION

As the primary way of sharing traffic information between controller and pilot in air traffic control (ATC) systems, voice conversation through radio transmission is very important to real-time traffic management. Limited by the technical issues, current ATC systems failed to detect speech errors from the voice conversation, which is further supposed to be a human-in-the-loop factor [1]. Recently, the workload of the ATC controller is continuously burdening with the increase of the civil aviation industry, which deteriorates the impact of human errors on flight safety. As investigated in [2], nearly 80% of ATC speeches do not strictly conform to the standard procedure, which may be regarded as an incentive to some real incidents. It is deeply received that learning situational context traffic dynamics from the ATC speech is essential to ensure operational safety, relieve the controller's workload and enrich the real-time traffic dynamics [3][4]. To this end, automatic speech recognition (ASR) is a promising technique to understand ATC speech, which is capable of bringing human factors into a closed ATC loop in an automatic manner.

As to the ASR research in the ATC domain, our previous works studied two independent end-to-end models for Chinese and English ASR tasks [3], cascaded pipelines for multilingual ASR task in a single controlling center [4] and its extension on majority conditions [5]. Note that all the aforementioned works were implemented with full supervised learning, i.e., optimize the ASR model by using the annotated data pairs (feature map, text sequence).

This work was supported by the National Natural Science Foundation of China (No: 62001315).
Correspondence: Y. Lin, College of Computer Science, Sichuan University, Chengdu 610000, China. (yilin@scu.edu.cn)

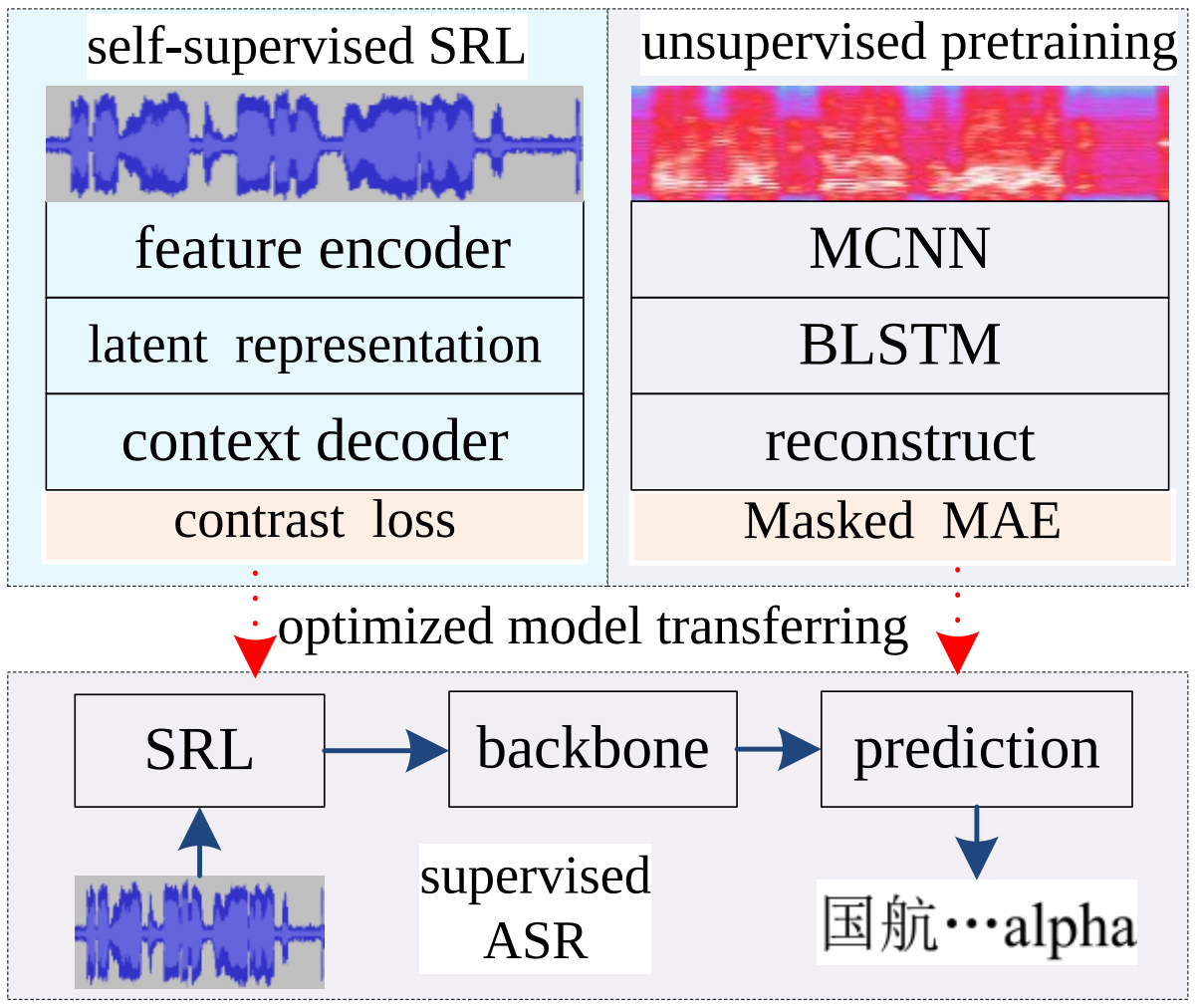

Fig. 1. The proposed framework.

It is widely demonstrated that a speech corpus, with a large data size, high speech diversity and vocabulary coverage, is a key to build a practical ASR model due to the intrinsic supervised learning mechanism [6]. For the common ASR research, various corpora have been built with different languages, in which up to thousand-hour speeches are available to train deep learning-based models, such as AISHELL-2 [7], Librispeech [8]. When it comes to the ASR research in the ATC domain, the speech corpus has several exclusive specificities, including the radiotelephony transmission, shared channel for multi-speakers, multilingual, terminologies, and unstable speech rate [9][10]. Technically, the following issues are raised to be considered in the ATC research.

1) Small transcribed samples: annotating ATC speech is a costly, laborious, and time-consuming work since the special terminologies, location-related words in the vocabulary are highly expertise-dependent. Only professional practitioners can be competent for this job. Thus, it is hard to build a qualified speech corpus for training a practical ASR system in the ATC domain. In short, the small sample problem is always an obstacle to the ASR research in the ATC domain.

2) Multilingual: as proved in [5], to advance the ASR technique to industrial ATC applications, studying the model on speeches in multiple languages is an inevitable problem that needs to be addressed. The multilingual issue in the ATC domain is also a special situation, wherein the speech for one instruction is spoken in multiple languages.

3) Feature representation: the challenges, i.e., radiotelephony speech, shared channel, multilingual and unstable speech rate, result in the fact that the ATC speech is with low signal-to-noise, inferior intelligibility compared to the common speech corpus. Naturally, there are reasons to believe that existing handcrafted feature engineering approaches may not be able to extract discriminative and robust features to meet the requirements of acoustic modeling for the ASR research.

Although great efforts have been made to cope with the mentioned challenges, it is also believed that a multilingual end-to-end framework with the ability of obtaining high performance on a small labeled corpus is a better solution for the ASR research in the ATC domain due to the elimination of cascaded errors and simply system architecture.

To this end, the learning mechanism is firstly proposed to characterize the speech features in the mentioned complex ATC environment. In this work, a speech representation learning (SRL) model is designed to capture discriminative and robust features from unlabeled waves by a **wave-to-feature** paradigm, in which the *self-supervised* learning strategy is applied to achieve the model optimization. A grapheme-based vocabulary is designed to integrate the multilingual ASR task into a single model, whose units are compatible with both Chinese and English speech. In the ASR model, an improved end-to-end architecture based on DeepSpeech2 (DS2) [11] is designed to serve as the backbone network. To address the issue of the small labelled ASR corpus in the ATC domain, an *unsupervised* pretraining strategy is proposed to optimize the backbone network with the feature of unlabeled speeches via **feature-to-feature** training. A mask prediction strategy is adopted to promote the model to learn both the temporal and frequency correlations from input features, which can be regarded as a unique parameter initialization for the final ASR learning. Finally, by combining the SRL, ASR backbone network and vocabulary, the proposed multilingual end-to-end ASR framework is achieved, named as ATCSpeechNet, which is trained on the transcribed speech corpus as a typical *supervised* learning task, i.e., **wave-to-text**. Basically, both the SRL and backbone pretraining aim to capture the high-level speech representations with weak labels that is able to be automatically obtained based on intrinsic characteristics of data samples. Most importantly, unlike existing methods, the ATCSpeechNet is able to directly translate the raw wave into multilingual human-readable text in a single model, without any feature engineering or lexicon.

The proposed framework is sketched in Fig.1. The SRL is firstly built to learn high-level abstract speech representations from raw audios from the perspective of feature learning. In the SRL network, a feature encoder is designed to map the raw audio to a latent space representation (LSR) which is further fed into the context decoder to generate speech features. Based on the intrinsic temporal characteristics of the speech wave, the self-supervised learning task is defined as: predict the next speech frames by considering the latent representations of previous inputs. Technically, the SRL is a sequential prediction task, whose input and output can be easily obtained from unlabeled speeches. The contrast loss is applied to evaluate the difference between the predicted speech frames and the real raw data.

Learning from existing works, an end-to-end based architecture is improved to serve as the backbone network of the ASR model, in which the convolutional neural network (CNN) and long short-term memory (LSTM) blocks are applied to mine the spatial and temporal correlations from the input feature. In general, the backbone network can also be regarded as a feature encoding module for the ASR task. To cope with the small transcribed sample problem, an unsupervised pretraining strategy is proposed to optimize the ASR backbone network from unlabeled speeches. To achieve the unsupervised training, a reconstruction module is designed to predict the masked speech features from the representations of the backbone network. The difference between the reconstructed feature and input feature is evaluated by a masked mean absolute error (MAE) loss function to update the trainable parameters of the backbone network. In this way, a dedicated parameter initialization is expected to be obtained from speech features to support the subsequent supervised optimization, concerning the final performance and the model convergence. Note that both the self-supervised SRL task and the unsupervised backbone pretraining task are performed on the unlabeled speech samples without any transcription.

By separately optimized with unlabeled speech samples, our proposed ASR framework is formulated by concatenating the SRL and backbone network, in which a prediction layer is designed to classify the extracted high-level features into the vocabulary words. The grapheme-based set with the Chinese characters and English letters serves as the acoustic modeling unit of the ASR model. In this work, the multilingual human-readable text sequence is the final representation of the proposed ASR model. The connectionist temporal classification (CTC) loss function [12] is applied to evaluate the training errors, and further upgrade the training parameters of the proposed model.

As a post process, a language model (LM) is also trained to improve the overall accuracy by correcting typical prediction errors from the perspective of semantics. In this work, the ATC terminologies allow the LM to learn their contextualized dependencies with common words, which is able to improve the performance of recognizing ATC entities. For example, the rules of the flight identification, i.e., a callsign with digits, are able to correct the acoustic prediction based on their contextualized information.

A large amount of real multilingual ATC speech (unlabeled) is applied to validate the proposed self-supervised and unsupervised learning, in which different area-related ATC terminologies, accents and flight phases (aerodrome, tower, approach, area, etc.) are concerned. The final ASR performance is evaluated on the ATCSpeech corpus (transcribed) [10]. The proposed ASR model achieves a 4.20% label error rate on only 58-hour transcribed samples, which obtains over 100% relative performance improvement on the baseline model. Notably, the improvement can be further promoted with the increasing of size of the transcribed samples. In addition, experimental results also confirm the effectiveness of different modules and training strategies in the proposed framework. In summary, this study contributes to the ATC related ASR research in five important ways:

1) A complete end-to-end framework is proposed to address the ASR issue, which has the ability of translating the raw wave into human-readable text directly, without any handcrafted feature engineering or lexicon.

2) A novel speech recognition paradigm is proposed to integrate the multilingual ASR task into a single model, which not only reduces the system complexity, but also obtains higher performance than that of the independent systems.

3) Facing the complex speech environment, the learning mechanism is proposed to characterize high-level speech features from unlabeled samples, instead of extracting speech features by well-designed handcrafted engineering methods. The speech representation learning mechanism is finally proved to be significant to improve the final ASR performance.

4) A three-stage training strategy, including self-supervised, unsupervised and supervised training, is proposed to cope with the problem of the small transcribed corpus in the ASR research, which allows the model to obtain superior performance even with a very small labeled corpus.

5) In this work, the full supervised learning mechanism of the ASR research is improved to solve the issues of data collection and annotation in the ATC domain. Based on the experimental results, we believe that the proposed approach is capable of advancing the ASR technique to industrial application in other vertical fields.

The rest of this paper is organized as follows. In Section 2, we introduce the works related to our research. The details of the proposed framework are introduced in Section 3. Section 4 lists the experiment configurations. Experimental results are reported and discussed in Section 5. The conclusions and future works are given in Section 6.

## II. RELATED WORKS

### *A. ASR related works*

ASR is an interdisciplinary research field that involves signal processing, pattern recognition, artificial intelligence, etc. The ASR research can be traced back to the 1950s [13], and has undergone several major technical improvements.

1) In the 1980s, the HMM/GMM framework [14][15] was the most popular ASR approach, in which the hidden Markov model (HMM) was applied to build the temporal transition among phoneme states [16]. In addition, the Gaussian mixture model (GMM) was proposed to fit the data distribution between speech frames and text labels. Almost the same time, the artificial neural network (ANN) based method was also proposed to solve the ASR task [17], which is also a fundamental component of current ASR models.

2) With the great development of the deep learning technique, the deep neural network (DNN) was widely applied to improve the GMM on modeling the data distribution [18], and hence formulated the HMM/DNN framework [19]. The HMM/DNN framework achieved higher performance than that of the HMM/GMM ones, and pushes the ASR research to the era of deep learning [20][21][22].

3) In the HMM-based ASR framework, the temporal correspondence between phoneme states and text labels is a required step for acoustic modeling, which is a laborious work and further limits the ASR performance. Therefore, Graves *et al.* proposed an end-to-end ASR paradigm [23], wherein a CTC loss function is able to automatically map the variable-length speech frames to variable-length text labels. The end-to-end framework not only showed desired performance superiority over other existing approaches, but also kept an intuitive pipeline from the speech feature to word sequence, directly [24][25]. Plenty of research outcomes have been obtained based on the end-to-end framework, such as the DS2 [11], CLDNN [26], FSMN [27], fully CNN [28], Jasper [29].

4) The sequence-to-sequence model was also explored to achieve the ASR task based on the end-to-end idea, such as the LAS [30]. The attention mechanism was further applied to improve the decoding performance by weighting the influence of speech features on text labels in a sequential task [31]. Lately, the Transformer based architecture was also introduced to address the ASR issues with the self-attention mechanism [32].

Traditionally, handcrafted feature engineering methods were widely applied to extract speech features from raw waves, which are further regarded as the input of existing ASR approaches, such as Linear Predictive Coding (LPC) [33], filterBank [34] and Mel Frequency Cepstrum Coefficient (MFCC) [35]. However, the handcrafted feature engineering method may not be optimal for all speeches. Fortunately, with the emergence of large-scale corpus and the computational improvement of hardware, it is precious practical to learn the speech features from raw waves, rather than obtained by handcrafted feature engineering methods [36]. The representation learning technique was studied to improve the performance of the ASR task [37]. The idea of wav2vec was explored to learn speech representations from waves in an unsupervised manner [38]. The ASR approaches that model the acoustic patterns on raw waves were also studied and achieve preliminary results [39][40][41].

As to the studies of multilingual ASR, a seq-to-seq based architecture was proposed to recognize several different Indian dialects [42]. The Unicode-based acoustic modeling unit was proposed to address the speech recognition and synthesis task on a Japanese and English joint corpus [43]. Similarly, a phoneme-based modeling unit was also studied to achieve the multilingual ASR task [44]. A network sharing approach was also developed to recognize the Chinese and English languages [45]. A multi-task learning mechanism was proposed to obtain an end-to-end multilingual task in [46].

### *B. ASR in ATC*

Currently, only minor corpora are available to the ASR research in the ATC domain. A multilingual (Chinese and English) corpus collected from a real ATC environment was released in our previous work, i.e., ATCSpeech [10]. Similarly, a simulated English ATC speech corpus was also published, i.e., ATCOSIM [47].

The specificities and challenges of the ASR research in the ATC domain were reviewed in [48]. In 2018, the Airbus company held an ASR challenge concerning speech recognition and callsign detection, and over 20 teams submitted their results [49]. A DNN based framework was proposed to study the speech recognition task on an English corpus with Czech and French accents [50]. A syntactic score-based decoding algorithm was also proposed to improve the ASR performance [51]. Srinivasamurthy *et al.* studied the ASR approach [52] in the ATC domain, in which the extracted prior knowledge is applied to improve the final performance. A semi-supervised approach was developed to achieve the iterative training of the ASR model [53].

Our previous work studied the monolingual end-to-end ASR model for Chinese and English, independently [3]. A cascaded pipeline was also proposed to address the multilingual ASR issues in the ATC domain [4]. Furthermore, a multiscale CNN architecture was explored to cope with the specificities for multilingual ASR task in the ATC domain [5].

## III. Methodologies

### *A. Speech representation learning*

Representation learning is a popular technique that focuses on capturing high-level and abstract features corresponding to the underlying causes or patterns of the observed data, with separate features or directions in feature space corresponding to different causes. It has been widely applied to train a universal encoder model that can be used for multiple down-stream tasks, which makes great contributions to many fields, such as NLP. The representation learning provides a compacting vector to depict salient hints for disentangling causal factors, which is very suitable for extracting the data features with complex and joint influencing patterns.

Facing the speech specificities in the ATC domain, the representation learning technique inspires us to reconsider the feature engineering for the ASR task. In this work, a speech representation learning model is proposed to provide more discriminative and robust features to support the acoustic modeling. Ideally, the SRL is expected to capture primary features that can be adapted to different specificities, including the radiotelephony speech, shared channel, multilingual and unstable speech rate, and discard

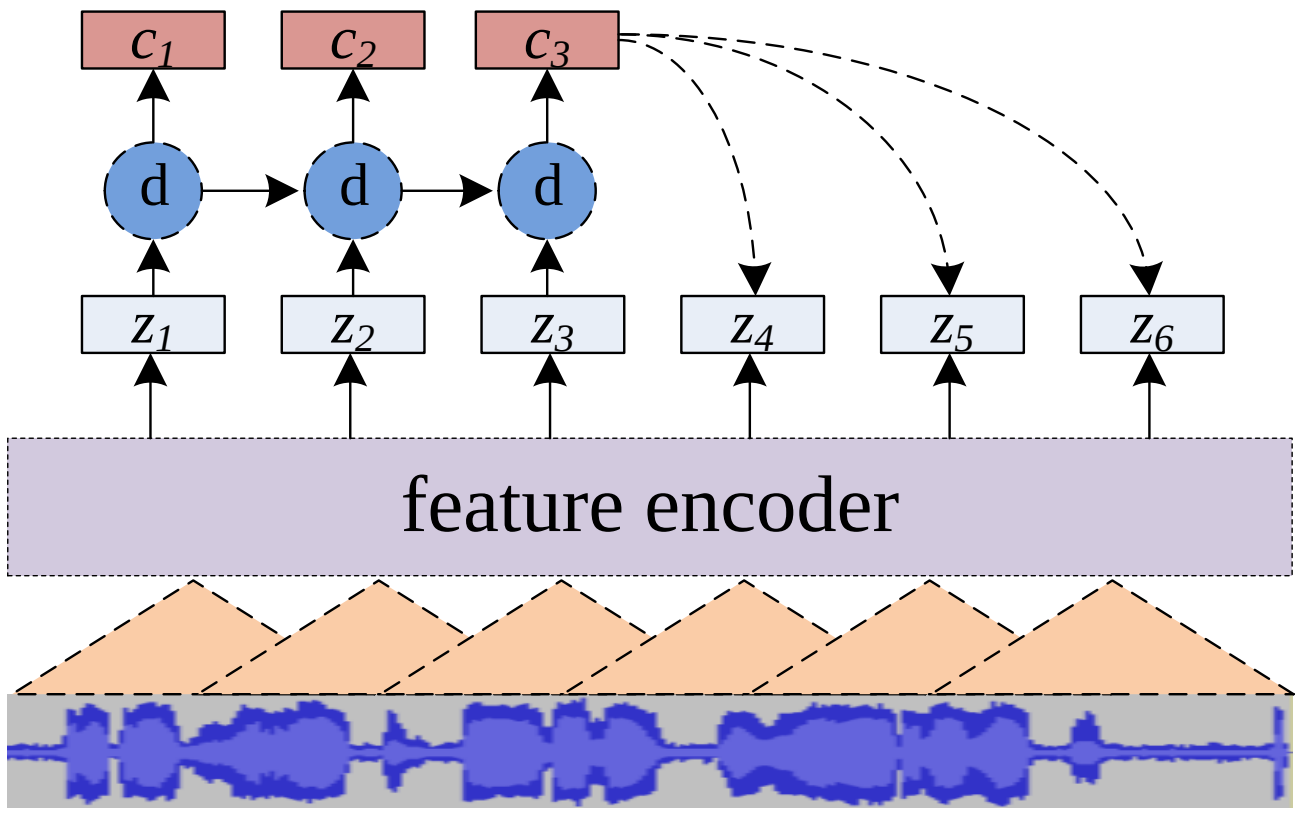

Fig. 2. The architecture of the SRL.

disturbing noisy features.

Learning from the state-of-the-art representation learning models, a self-supervised learning mechanism is applied to obtain the speech representations without any text transcription. To achieve self-supervised learning, the training task is defined as: predicting the next speech frames from previous input frames. The training samples of the self-supervised task can be automatically extracted from raw waves due to their intrinsic temporal characteristics. In this work, since the speech wave is one-dimensional data with a high sample rate, we first propose to map them to a lower and more compact temporal dimension and high feature dimension space by the neural network activation, i.e., LSR. The LSR mainly mines the frame-wisely speech feature, which is further applied to generate the contextualized speech features by considering the frame dependencies.

The whole SRL paradigm is shown in Fig.2. The raw speech wave ( $X$ ) is fed into a feature encoder to generate the speech representation in a latent space, i.e., $z_t$ . In this work, a CNN architecture is designed to serve as the feature encoder, as (1), $f_{enc}$ denotes the inference rule of the CNN blocks. Different CNN layers take different local receptive fields as a modeling unit, which aims to cope with the specificities of ATC speech, including acoustic granularity of different languages and the unstable speech rate. Based on the LSR, a context decoder is applied to capture the contextualized information among the whole input sequence, i.e., $c_t$ . As shown in (2), $x_t \in X$ is the input speech, and $f_{dec}$ is the inference of the context vector. In this work, the context decoder is also a CNN network, which considers its past context outputs to predict the next multiple frames.

$$z_t = f_{enc}(x_t) \quad (1)$$

$$c_t = f_{dec}(z_{\le t}) \quad (2)$$

Based on the idea of the instance discrimination, the contrast loss is applied to train the model by distinguishing whether the predicted $k$ step $z_{t+k}$ is drawn from a proposal distribution [54], as shown below in (3) and (4). $T$ is the total sequence length, and $\sigma$ is the sigmoid operation. $\tilde{z}$ is drawn from the proposal distribution, such as $T^{-1}$ [38], and $\lambda$ is the number of samples. $h$ is a fully connected neural network.

$$z_{t+k} = SRL(x_{t+k}, c_t) \quad (3)$$

$$L = -\sum_{s=1}^{k}\sum_{i=1}^{T-s}\left(\log\sigma(z_{i+s}, c_i) + \lambda \mathrm{E}\left[\log\sigma(-\tilde{z}, c_i)\right]\right) \quad (4)$$

$$\sigma(z,c) = \frac{1}{1+\exp(-z\cdot h(c))} \quad (5)$$

Note that, in the proposed network, both the $z$ and $c$ can be regarded as a representation of the input speech wave, in which they focus on the framewise and contextualized (sequential) features, respectively. In this work, as a typical sequential classification task, the ASR model tends to take the contextualized representations to make a globally optimal prediction, i.e., $c$ .

## *B. Backbone network pretraining*

For the ASR research, a long-standing idea is that transcribed speech samples take substantial efforts to obtain a high model performance. However, in the ATC domain, building a qualified transcribed corpus is costly and time-consuming. It is urgent and worthwhile to explore how to effectively utilize the unlabeled ATC speech to obtain desired accuracy even with a small labeled data. In this work, a pretraining strategy is proposed to perform a preferred warmup optimization on the backbone network from the speech features of the unlabeled samples. In this work, an end-to-end architecture is improved to serve as the backbone network based on our previous work [5], in which the several CNN and LSTM layers are applied to build the frequency and temporal dependencies, respectively. Similarly, the multi-scale CNN (MCNN) architecture is also designed to capture the high-level abstract features in different level, which aims to address the diverse data distribution caused by the volatile background noise model, multilingual speech, unstable speech rate, etc.

In order to achieve the unsupervised training for the backbone network, a reconstruction module is designed to recover the speech features by the transposed convolutional layer. The reconstruction module aims to generate the speech feature whose size is same with that of the model input. The reconstructed features are applied to measure the prediction errors and further upgrade the trainable parameters of the backbone network. The feature of unlabeled speeches serves as both the input and output of the network.

Learning from the idea of denoising networks, a mask prediction strategy is proposed to ensure that the pretraining promotes the model to learn the data patterns, rather than copy. In the input speech feature, 15% of the frames are randomly selected to perform the masking strategy. Drawn a random number $p$ from Equidistribution (0 to 1), the selected frames $f_t$ are converted into $\tilde{f}_t$ in the following three ways:

1) If $p < 0.1$, $f_t$ is set to zero vector, i.e., $\tilde{f}_t = \vec{0}$.

2) If $0.1 < p \le 0.2$, $f_t$ is set to a vector drawn from a normal distribution with the same shape of $f_t$.

3) Otherwise, $\tilde{f}_t$ is generated based on its temporal neighbors, as (6). Considering the duration of an acoustic unit in the ATC domain [5], $N$ is set to 5 in this work.

$$\tilde{f}_t = \frac{1}{2N}\sum_{i=-N}^{N} f_i \qquad (6)$$

The target of the training is to minimize the L1 loss between the reconstructed frames and their ground truth:

$$\arg\min_{\theta}\left(\frac{1}{M}\sum_{i=1}^{M} m_i \left|F_i - \tilde{F}_i\right|\right) \qquad (7)$$

where $M$ is the mini-batch size. $F_i$ and $\tilde{F}_i$ are the real and predicted speech features for *i-th* sample, respectively. $m_i$ is the mask vector, whose length is same with the frame number $T_i$. In the mask vector, $m_i^t, t \in [0, T_i]$ is set to 1 if the *t-th* frame is selected to be performed the masking strategy, otherwise, it is 0. The mask vector is to ensure that only the masked frames are considered to compute the prediction error, and further achieve an effective back-propagation.

$$m_i = \left[m_i^1, m_i^2, \cdots, m_i^{T_i}\right] \qquad (8)$$

With the proposed pretraining procedure, the backbone network is expected to learn the speech representations from unlabeled speeches, and obtain a preferred initialization for its trainable parameters. The initialization is applied to speed up the model convergence on the transcribed samples, and further improve the final ASR performance.

### *C. The proposed ASR model*

In general, both the self-supervised speech representation learning and unsupervised pretraining of the backbone network are to learn significant patterns from unlabeled speeches. In this section, the supervised training is performed to accomplish the final ASR task, i.e., text prediction.

To integrate the multilingual ASR into the same framework, we design a compatible vocabulary to serve as the modeling unit of the ASR model from the perspective of grapheme. The vocabulary comprises Chinese characters, English letters, and several special tokens (i.e., "<SPACE>", "<UNK>" and the apostrophe).

To achieve an end-to-end ASR framework, the SRL network is firstly combined with the backbone network the obtain desired speech representations from the perspective of feature learning and pretraining. In succession, a fully connected layer is designed to serve as the predictor based on the learned representations. The time-distributed mechanism is performed on the temporal dimension to achieve the ASR task (i.e., sequential classification). The architecture of the proposed end-to-end multilingual ASR model is depicted in Fig. 3. The output of the model is the probability of given words condition on the input speech frame-wisely. The output shape is $(T, |V|)$, in which $T$ is the number of the frames, $V$ and $|V|$ are the vocabulary and its size, respectively.

The errors between the prediction and the real label are computed by the CTC loss function. Let the input is $F = \langle f_1, \cdots, f_T \rangle$ and $y_k^t$ denote the probability that the *t-th* frame corresponds to the output label $k$. For a certain input speech, the probability of an output sequence $\pi$ is measured as (9). The probability of the final text sequence $l$ is computed by summing all the possible ones denoting by $\Xi$. For example, using '_' to denote blank, both the output 'a_bb__c' and '_a_b_c_' are corresponding to the final text "abc".

$$p(\pi \mid F) = \prod_{t=1}^{T} y_{\pi_t}^t, \pi_t \in V \qquad (9)$$

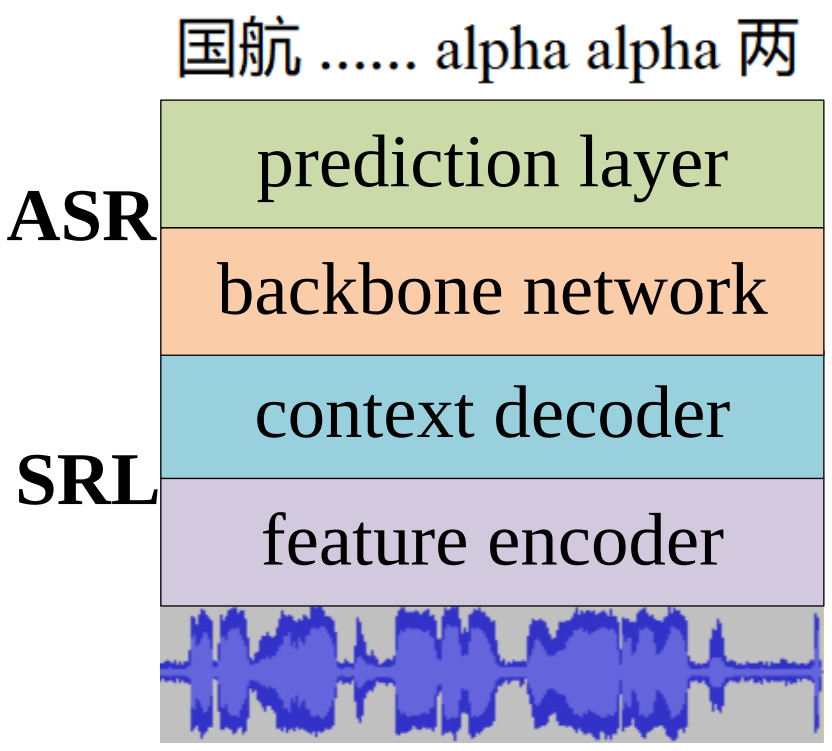

Fig. 3. The whole paradigm of the proposed ASR model.

$$p(l\,|\,F)=\sum_{\pi\in\Xi^{-1}(l)} p(\pi\,|\,F) \tag{10}$$

## D. *Training strategies*

Typically, the SRL can be regarded as a speech feature extractor. During the ASR training, the SRL model is to convert the raw wave into learned features which further serve as the input of the ASR model. Obviously, the two models are trained on their own loss functions, i.e., contrast loss and CTC loss. To unify the learning target, an integrated optimization procedure is proposed to train the ASR model, where both the parameters of the SRL and ASR model are upgraded, simultaneously. Consequently, a three-stage training strategy is formulated to train the proposed end-to-end ASR model, from speech wave to texts, as shown below:

a) representation learning: the SRL model is trained on the unlabeled corpus with the self-supervised learning mechanism.

b) backbone pretraining: the backbone network of the proposed ASR model is pre-trained on the speech features in an unsupervised manner, in which the features are generated by the SRL model from the unlabeled corpus.

c) integrated optimization: the whole network, including the SRL, backbone, and the prediction layers, is trained on a transcribed speech corpus, as the typical supervised learning ASR task.

In general, stage a) and b) can be achieved without any speech transcription, which focuses on learning the universal patterns from the raw speeches. In stage c), the model training is performed on a transcribed corpus, which aims to capture the probability distribution between the speech feature and the text label.

Obviously, the proposed three-stage training strategy is capable of addressing the annotation difficulties of the ATC related speech corpus. In addition, the proposed SRL model not only captures more robust and discriminative features to support the acoustic modeling, but also greatly speeds up the convergence progress even in a small labeled training corpus.

TABLE I
DESCRIPTIONS OF UNLABELED SPEECH CORPUS

| Language | #hours | #utterance | #area | #flight phase |
|---|---|---|---|---|
| Chinese | 2059 | 2361528 | 12 | 4 |
| English | 556 | 510073 | 12 | 4 |

TABLE II
EXPERIMENTAL DESIGN AND MODELS

| ID | Model | | Training strategy | |
|---|---|---|---|---|
| | SRL | ASR | backbone pretraining | integrated training |
| A1 | | ASR-HC | | |
| A2 | / | ASR-HE | No | No |
| A3 | | ASR-HA | | |
| B1 | SRL-C | ASR-LC | | |
| B2 | SRL-E | ASR-LE | No | No |
| B3 | SRL-A | ASR-LA | | |
| C1 | SRL-A | ASR-LA | Yes | No |
| C2 | SRL-A | ASR-LA | No | Yes |
| C3 | SRL-A | ASR-LA | Yes | Yes |
| C4 | SRL-A | ASR-LAX | Yes | Yes |

# IV. EXPERIMENT CONFIGURATIONS

## A. *Data description*

The raw speeches were collected from the real ATC systems in China, such as Chengdu, Chongqing, Zhengzhou, Shanghai, Kunming, Shenyang, *et al*. The raw speeches were saved as one file per hour continuously, which means that both the silence and noise were also recorded. The raw speech is firstly split into segments to simulate the real controlling scene by the voice activity detection operation, and the silence and noise are removed from the speech segments.

In this work, several corpora are built to validate the proposed framework. The unlabeled speech corpus is applied to train the SRL and the backbone network, which is detailly described in Tab. 1, concerning the duration, the number of utterances, areas and flight phases. The unlabeled corpus is called as **RUD**, in which the Chinese and English ones are named as **RUD-C** and **RUD-E**, respectively. The training set of our released transcribed speech corpus, i.e., **ATCSpeech**, is applied to optimize the proposed approach in the supervised optimization phase, as described in [10]. The Chinese and English speeches in ATCSpeech are named as **ATCSpeech-C, ATCSpeech-E**, respectively.

Furthermore, a larger corpus (**ATCSpeech-large**) is also applied to confirm the performance improvement obtained by increasing the labeled training corpus. The duration of the Chinese and English speech samples in ATCSpeech-large are about 79 and 36 hours, respectively, i.e., nearly a doubled data size of the ATCSpeech. The vocabulary size increases from 679 to 730 due to the new transcribed samples. Note that all the speeches of the ATCSpeech corpus are entirely included in the ATCSpeech-large corpus.

In this work, the test dataset in ATCSpeech is applied to evaluate the final performance of the ASR model, called as **TTC**. The Chinese and English speeches in TTC are named as **TTC-C, TTC-E**, respectively. To keep the fairness, all the experiments are evaluated on the TTC dataset and the same dataset division is applied to all the designed experiments.

The transcriptions in the labeled corpus (ATCSpeech) are further used to train the character-level LM for improving the final performance. In this work, the classical n-gram LM [55] is trained to implement an n-best decoding strategy [11], in which the $\alpha$ and $\beta$ are set to 1.25 and 1.5, respectively.

In this work, the final performance is measured by the label error rate (LER), i.e., word error rate based on the Chinese characters and English words, as shown below:

$$LER = \frac{(I + D + S)}{N_t} \quad (11)$$

where $N_t$ is the length of the ground-truth, while the $I$ , $D$ and $S$ are the number of the insert, delete and substitution operations for converting the predicted label into the ground-truth.

*B. Experimental Design*

To validate the proposed ASR approach in this work, several experiments are designed to validate the multilingual ASR model, representation learning, pretraining and integrated optimization. The details of the designed experiments are summarized in Tab. 2. In the listed items, the backbone pretraining indicates that the backbone network of the ASR model is pre-trained on the RUD dataset before its optimization on the ATCSpeech corpus. The integrated optimization indicates that the SRL network is also optimized during the supervised training of the ASR model. To achieve the designed experiments, a total of 3 SRL models are trained, as described below:

a) SRL-C: the model is trained on the RUD-C dataset.
b) SRL-E: the model is trained on the RUD-E dataset.
c) SRL-A: the model is trained on the RUD dataset.

Similarly, the following ASR models are required to conduct the designed experiments, as shown below:

a) ASR-HC: the model is optimized on the ATCSpeech-C with handcrafted filterBank features.
b) ASR-HE: the model is optimized on the ATCSpeech-E with handcrafted filterBank features.
c) ASR-HA: the model is optimized on the ATCSpeech with handcrafted filterBank features.
d) ASR-LC: the model is optimized on the ATCSpeech with learned features generated by the SRL-C.
e) ASR-LE: the model is optimized on the ATCSpeech with learned features generated by the SRL-E.
f) ASR-LA: the model is optimized on the ATCSpeech with learned features generated by the SRL-A.
g) ASR-LAX: the model is optimized on the new ATCSpeech-large corpus, whose input is the learned features generated by the SRL-A.

Note that all aforementioned models (SRL and ASR) are built with the same architecture. The number of neurons in the prediction layer is set to the size of the vocabulary.

In the designed experiments, the experiments in ‘group A’ are developed with the common approach in existing works, i.e., handcrafted feature engineering and full supervised learning, which also serve as the baseline in this work [10]. The experiment for proving the performance improvement of multilingual ASR task is also designed in this group. The experiments in ‘group B’ aims to validate the proposed speech representation learning mechanism, while the ‘group C’ focuses on confirming the efficient and effectiveness of the proposed training strategies, including the pretraining and integrated optimization. Finally, experiment C4 is also designed to consider the advantages by increasing the size of the labeled corpus.

*C. Training Configurations*

In this work, deep learning models are constructed based on the open framework Pytorch 1.1.0. The training server is configured as follows: 2*Intel Core i7-6800K, 2*NVIDIA GeForce RTX 2080Ti and 64 GB memory with operation system Ubuntu 16.04.

In the SRL training, the Adam optimizer with $10^{-6}$ initial learning rate is applied to optimize the weight parameters. A learning rate scheduler [38] based on the cosine rule is applied to dynamically adjust the learning rate during the whole training, whose minimal and maximal learning rate are set to $10^{-9}$ and $10^{-3}$, respectively. The warmup strategy is applied to improve the training stability, in which the warmup iteration and learning rate are 500 and $10^{-7}$, respectively. In addition, the number of negative samples for the contrast loss is 10. All the samples are concatenated and divided into about 10 seconds, which serve as the input of the SRL model.

In the backbone network pretraining, the Adam optimizer with $10^{-4}$ initial learning rate is applied to optimize the weight parameters. The learning rate will be dynamically halved every 25 training epochs. The batch size is 64. In the temporal dimension, the samples are padded to meet the requirement that the number of frames is even, so that the shape of the final output is the same as that of the input features.

TABLE III
PERFORMANCE OF MULTILINGUAL ASR TASK

| Id | TTC-C (%) | | TTC-E (%) | | TTC | |
|---|---|---|---|---|---|---|
| | Greedy | LM | Greedy | LM | Greedy | LM |
| A1 | 8.10 | 6.31 | - | - | - | - |
| A2 | - | - | 10.40 | 9.20 | - | - |
| A3 | **6.96** | **5.95** | **9.04** | **7.80** | 8.09 | 6.96 |

TABLE IV
PERFORMANCE OF ASR MODEL WITH SRL

| Id | TTC-C (%) | | TTC-E (%) | | TTC | |
|---|---|---|---|---|---|---|
| | Greedy | LM | Greedy | LM | Greedy | LM |
| B1 | 4.13 | 3.59 | - | - | - | - |
| B2 | - | - | 6.01 | 5.58 | - | - |
| B3 | 3.95 | 3.47 | 5.74 | 5.33 | 5.11 | 4.90 |

During the optimization of the ASR model, the Adam optimizer is used to update the weight parameters. The initial learning rate is set to $10^{-4}$ and it will be halved every 25 training epochs. Since both the SRL and ASR models are trained with their own loss evaluation, the warmup strategy is also applied to keep a smooth updating during the integrated optimization. The warmup iteration and learning rate are 1000 and $10^{-5}$, respectively. The batch size is set to 128 for two GPUs parallel training. In order to reduce the loss to a certain level as soon as possible, all training samples are sorted by their durations in the first training epoch. In ATC speeches, a similar duration of training samples may indicate that their texts belong to the same controlling instruction and share higher similarity among them. All training samples are shuffled after the first training epoch to improve the robustness of the model. An early stopping strategy based on the validation loss is applied to check the training progress.

## V. RESULTS AND DISCUSSIONS

### A. Multilingual ASR

Basically, the experiments in group A focus on confirming the performance improvement of the multilingual ASR training in the ATC domain. The experimental results are reported in Tab. 3, in which both the LERs obtained by greedy decoding and LM decoding are listed.

It can be seen from the experimental results that the multilingual ASR training benefits the final performance on the ATC related corpus. In the multilingual training, both the Chinese and English speech improve over 1% absolutely LER for the acoustic modeling. As to the LM decoding, the final LER is further reduced, about 0.4% and 1.5% absolutely LER improvement for Chinese and English speech, respectively. The LM for English speech obtains higher improvement since the modeling unit (English letter) is more basic and able to provide more intensive contextualized dependencies among them.

In addition, we also consider the language classification task for the multilingual issue in the ATC domain. Experiment results show that it is also achieved with considerable high accuracy, up to 99.90% with LM decoding. Some classification errors still occur due to the ATC specificities, i.e., multilingual texts in a single utterance.

In summary, for the ASR research in the ATC domain, the multilingual training with the proposed vocabulary benefits the performance improvement in an end-to-end manner. The proposed approach not only obtains desired performance superiority, but also formulates a multilingual end-to-end ASR framework with a simpler system pipeline. The most prominent significance is that the multilingual ASR allows us to be free from the language classification task in practice. For the independent ASR systems, the language classification task is prone to fall into the cascaded prediction error with the ASR task. If the speech is predicted as an error language, the ASR texts will be completely mistaken.

### B. Representation Learning for ASR

In this section, the experiments in group B aim to prove the effectiveness and efficiency of the representation learning on advancing the ASR performance in the ATC domain. In this section, the independent ASR models for monolingual are also trained to clarify the improvements of the multilingual and the SRL, separately. The experimental results are listed in Tab. 4.

From the experimental results, we can see that the proposed representation learning mechanism makes great contributions to advance the final ASR performance, even with monolingual training and a very small transcribed corpus. The results can be summarized as listed below:

a) On one hand, the SRL significantly improves the LER with the same model architecture on the same training corpus. For Chinese speech, the LER of the greedy decoding is improved from 8.10 to 4.13, while it is from 10.40 to 6.01 for English speech. That is to say, the SRL is capable of obtaining about 96% and 73% relative LER improvements even with monolingual training using the same model architecture. In terms of improving the final performance, it can be concluded that the SRL contributes more

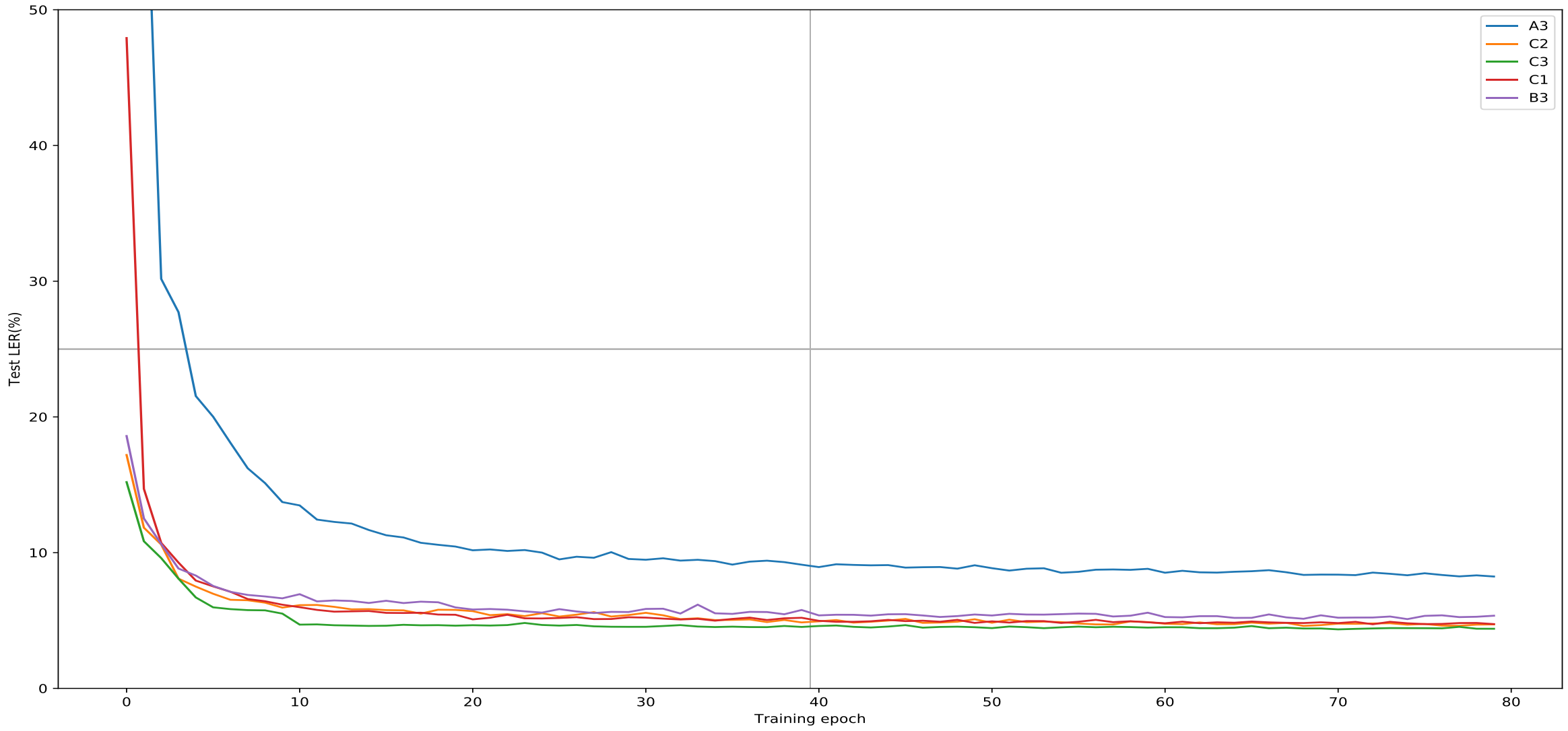

Fig. 4. The comparison of the test LERs.

than the multilingual training.

b) On the other hand, the SRL allows the multilingual ASR model to yield an excellent prediction accuracy with a very small transcribed corpus (only 36-hour Chinese and 18-hour English speech). The final LER is only 4.90%, which achieves a 42% relative improvement compared to that of the handcrafted feature engineering. It can be concluded that the SRL is able to advance the ASR research in the ATC domain to an industrial scale, i.e., less than 5% LER.

c) With the application of the SRL, universal data patterns are extracted from unlabeled speech, which encourages a high-performance acoustic modeling between learned speech features and text labels in the supervised learning phase, as well as the language classification. As demonstrated in the results, compared to the handcrafted features, the LM decoding for SRL provides inferior performance improvement, about 1.2% vs. only 0.2% absolutely LER reduction. In addition, the language classification accuracy is further improved to 99.97%.

d) Finally, the SRL-driven acoustic modeling is extremely helpful to speed up the model convergence. As shown in Fig.4, the test LER of experiment B3 (SRL and multilingual) is rapidly dropped to about 10% in the third training epoch, when the LER of experiment A3 (handcrafted feature engineering and multilingual) is still about nearly 30%.

In conclusion, the proposed SRL boosts the performance from the perspective of the data size of the transcribed corpus (reduces the annotation cost) and the progress of the model convergence. Specifically, the proposed SRL is expected to be a decisive factor for addressing the small training corpus in the ATC domain by learning high-level speech features by taking advantage of a large number of unlabeled speeches.

### *C. Training strategies and data size*

In last the two group experiments, the multilingual ASR and SRL have been shown to be effective ways to improve the final ASR performance. Therefore, both the multilingual task and representation learning are directly applied to the experiment in group C. The purpose of this group is to validate the proposed training strategy, including the unsupervised backbone network pretraining and the integrated supervised optimization. The experimental results are reported in Tab. 5, in which the LER is evaluated on the TTC only.

It can be seen from the experimental results (C1 and C2) that both the pretraining and the integrated optimization have the ability of improving the ASR performance. The final LER is slightly improved to 4.75% and 4.87%, compared to the results of experiment B3. By applying both the backbone pretraining and the integrated optimization (C3), the final LER is further improved to 4.20%.

In this group, a larger ATCSpeech corpus is also used to train the ASR model for the supervised optimization, which aims to demonstrate the influence of data size on the final ASR performance. As shown in the results (C4), the final measurement is further improved by increasing the size of the transcribed training corpus (about 115-hours), from 4.20% to 3.35%. The final LER is expected to be improved by obtaining more transcribed speech samples. All in all, for the ASR task, the 115-hours corpus is not able to be regarded as large enough.

To further clarify the influence of the proposed approach on the training process, the test LERs for different experiments with the elapse of the training epoch are visualized in Fig.4. From the experimental results we can obtain the following summaries:

a) Both the proposed SRL, ASR model and their training strategies are proved to be worked for improving the ASR performance in the ATC domain.

b) By comparing the results of A3 with others, it can be concluded that the SRL makes incredible contributions to advance the final performance.

TABLE V
PERFORMANCE OF THE ASR MODEL WITH DIFFERENT STRATEGIES

| Id | Greedy (%) | LM (%) |
|---|---|---|
| C1 | 4.98 | 4.75 |
| C2 | 5.01 | 4.87 |
| C3 | 4.61 | 4.20 |
| C4 | 3.77 | 3.35 |

TABLE VI
PERFORMANCE OF GENERALIZATION TESTS

| | decode | D1 | D2 |
|---|---|---|---|
| CN LER (%) | Greedy | 17.15 | **13.63** |
| | LM | 12.21 | **10.16** |
| EN LER (%) | Greedy | 10.70 | **8.10** |
| | LM | 8.55 | **6.53** |
| Language Acc. (%) | Greedy | 99.94 | **100.00** |
| | LM | 99.96 | **100.00** |

c) As shown in the results, the LER of the first epoch for the experiment A3 and B3 are over 40%, while other experiments obtain less than 20% LER. The improvement can be attributed to the proposed training strategies, in which a fine-tuned parameter initialization is learned from unlabeled speeches.

d) The proposed ASR approach and training strategies are capable of obtaining expected performance superiority even with a very small transcribed corpus in a multilingual and end-to-end manner, which is significant to advance the ASR technique into industrial applications.

### *D. Generalization Test*

In this section, two open corpora are also used to validate the generalization of the proposed approach. A total of 1860-hour corpus called **GTP**, including AISHELL2 [7] and Librispeech (train-clean-360 and train-other-500) [8], serves as the unlabeled corpus for the SRL training. A joint corpus (AISHELL-1 [56] and the train-clean-100 in Librispeech), called as **GTS**, serve as the transcribed corpus of the ASR optimization, about 200 hours. Finally, the results of the proposed approach on open corpora are evaluated on the test-clean-100 of Librispeech (English) and the test division in AISHELL-1 (Chinese). The experiments are described below:

1) D1: The model is trained on the GTS with the handcrafted features (log filter Bank), whose configurations are same with that of in experiment A3.

2) D2: The model is trained on the GTS with the proposed training strategies, in which the SRL model is optimized on the GTP dataset. The configurations are same with that of in experiment C3.

The results of the generalization test are reported in Tab. 6. In general, although all the strategies are proposed to address the ATC related issues in this work, they are also worked for common ASR corpus. With the proposed strategies, both the LER and the language classification are improved in experiment D2.

Due to the high speech distinction, i.e., rigorous monolingual in one utterance, the language classification task is achieved without any prediction error. The accuracy of the acoustic modeling is also evidently improved, absolutely 3.5% and 2.6% LER reduction for Chinese and English speech, respectively. After applying the LM decoding, the final LERs for Chinese and English speech are 10.16% and 6.53%, respectively, with only 100-hour transcribed samples.

It can also be seen that, compared to the experimental results with that of in the ATC domain, the open corpora obtain fewer benefits of the performance improvement. It is indicated that the handcrafted feature engineering method is already able to extract qualified speech features from stationary speeches (near-field reading speeches) in the open corpora, which lays a solid foundation for acoustic modeling. Thus, we come to the conclusion that the proposed approach is able to contribute more to the ASR research with the following specificities: complex speech features (volatile noise, multi-speakers, multilingual, etc.), multilingual end-to-end paradigm, and a very small transcribed corpus.

## VI. CONCLUSIONS AND FUTURE WORK

In this work, to achieve the speech recognition task in the ATC domain, a complete multilingual end-to-end ASR model is proposed which is capable of translating the raw wave into text in a single step, without any feature engineering or lexicon. Facing the challenges of volatile background noise and inferior intelligibility, unstable speech rate, multilingual speech, the learning mechanism is firstly proposed to extract robust and discriminative features from raw waves to support the acoustic modeling. A CTC-based end-to-end model is improved to serve as the acoustic model, in which a compatible vocabulary, including Chinese characters and English letters, is designed to cope with the multilingual ASR task. A three-stage training strategy is proposed to deal with the issue of the small transcribed corpus in the ATC domain, including the self-supervised representation learning, backbone ASR network pretraining, and integrated optimization. The LMs are also trained to correct spelling errors, separately, from the perspective of semantics. Experimental results on the ATCSpeech corpus demonstrate that the proposed approach is able to achieve an excellent performance on a very small corpus, i.e., 4.20% LER on only a 54-hours transcribed corpus. It also shows that the performance will be further improved by increasing the transcribed training corpus. Furthermore, the proposed ASR approach and training strategy is also proved to be worked on common ASR corpora, such as AISHELL and Librispeech. In conclusion, the proposed approach not only obtains desired performance improvement (almost industrial level), but also reduces the system complexity for the multilingual ASR task. The proposed ASR framework can serve as strong support of air traffic systems, such as understanding the controlling intent from the real-time controlling scene.

In the future, we first plan to increase the size of the transcribed training corpus to cover more words and improve the data diversity. More effective network architectures, such as the attention mechanism, Transformer, et al., are also deserved to be studied to address this issue. Last but not least, it is also important to advance the ASR technique to real ATC related applications.